# Multiclass Road Sign Detection using Multiplicative Kernel

Valentina Zadrija
Mireo d. d.
Zagreb, Croatia
valentina.zadrija@mireo.hr

Siniša Šegvić
Faculty of Electrical Engineering and Computing
University of Zagreb
Zagreb, Croatia
sinisa.segvic@fer.hr

*Abstract*—We consider the problem of multiclass road sign detection using a classification function with multiplicative kernel comprised from two kernels. We show that problems of detection and within-foreground classification can be jointly solved by using one kernel to measure object-background differences and another one to account for within-class variations. The main idea behind this approach is that road signs from different foreground variations can share features that discriminate them from backgrounds. The classification function training is accomplished using SVM, thus feature sharing is obtained through support vector sharing. Training yields a family of linear detectors, where each detector corresponds to a specific foreground training sample. The redundancy among detectors is alleviated using *k*-medoids clustering. Finally, we report detection and classification results on a set of road sign images obtained from a camera on a moving vehicle.

*Keywords—multiclass object detection; object classification; road sign; SVM; multiplicative kernel; feature sharing; clustering*

## I. Introduction

Road sign detection and classification is an exciting field of computer vision. There are various applications of the road sign detection and classification in driving assistance systems, autonomous intelligent vehicles and automated traffic inventories. The latter one is of particular interest to us since traffic inventories include periodical on-site assessment carried out by trained safety expert teams. Current road condition is then manually compared against the reference state in the inventory. In current practice, the process is tedious, error prone and costly in terms of expert time. Recently, there have been several attempts to at least partially automate the process [1], [2]. This paper presents an attempt to partially automate this process in terms of road sign detection and classification using the multiplicative kernel.

In general, detection and classification are typically two separate processes. The most common detection method is sliding window approach, where each window in the original image is evaluated by a binary classifier in order to determine whether the window contains an object. When the prospective object locations are known, classification stage is applied only on the resulting windows. One common classification approach includes partitioning the object space into subclasses and then training a dedicated classifier for each subclass. This approach is known as one-versus-all classification.

In this paper, we focus on detection and classification of ideogram-based road signs as a one-stage process. We employ a jointly learned family of linear detectors obtained through Support Vector Machine (SVM) learning with multiplicative kernel as presented in [3]. In its original form, SVM is a binary classifier, but multiplicative kernel formulation enables the multiclass detection. Multiplicative kernel is defined as a product of two kernels, namely between-class kernel $k_x$ and within-class kernel $k_\theta$ as described in Section IV. Between-class kernel $k_x$ is dedicated for detection, i.e. the foreground-background classification. Within-class kernel $k_\theta$ is used for within-foreground classification, i.e. to discriminate between various object subclasses. The result of SVM training is a set of support vectors and corresponding weights which are then used to generate a family of detectors. The detectors are obtained by tuning within-class state into the multiplicative kernel as described in Section IV-B. The key point here is that all detectors share the same support vectors, but weights of specific support vectors vary depending on the within-class state value.

We evaluate the described approach on a set of predefined road sign subclasses defined in Section III and present experimental results in Section V.

## II. Related Work

In recent years, a lot of work has been proposed in the area of multiclass object detection. We will address the issues from both road sign detection aspect as well as general multiclass object detection and feature sharing.

The approach presented in [4] tries to solve multi-view face detection problem. Foreground class is partitioned into subclasses according to variations in face orientation. For each subclass, a corresponding detector is learned. This approach exhibits several problems when used with a larger number of subclasses. More specifically, number of features, number of false positives and the total training time grow linearly with the number of subclasses.

This work has been supported by research projects Vista (EuropeAid/131920/M/ACT/HR) and Research Centre for Advanced Cooperative Systems (EU FP7 #285939).

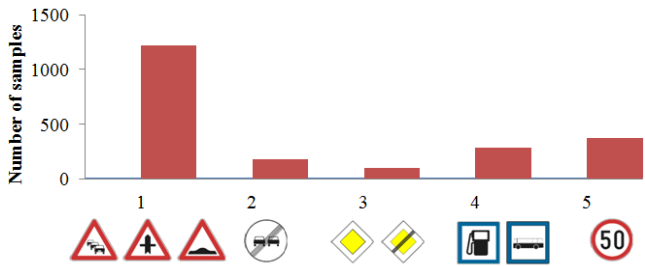

Fig. 1. Distribution of samples with respect to road sign subclasses $S = \{1, 2, 3, 4, 5\}$ for training dataset. Below each subclass label $v$, representative subclass members are shown. Road signs shown for subclass $v=1$ are informal, i.e. subclass contains 9 different road signs.

Further, authors in [5] focus on feature sharing using JointBoost procedure. In contrast to [4], where number of features grows linearly with the number of subclasses, the authors have experimentally shown that the number of features grows logarithmically with respect to the number of subclasses. Additionally, the authors showed that a jointly trained detector requires a significantly smaller number of features in order to achieve the same performance as a independent detector.

The approach presented in [6] deals with problem of multi-view detection using the so called Vector Boosted Tree procedure. At the detection time, the input is examined by a sequence of nodes starting from the root node. If the input is classified by the detector of current node as a member of the object class, it is passed to the node children. Otherwise, it is rejected as a background. The drawback of this approach is that it requires the user to predefine the tree structure and choose the position to split.

Similar to [6], the approach presented in [7] also employs a classifier structured in a form of a tree. However, in contrast to [6], the tree classifier is constructed automatically - the node splits are achieved through unsupervised clustering. The algorithm is iterative, i.e. at the beginning it starts with an empty tree and training samples which are assigned weights. By adding a node to the tree, the sample weights are modified accordingly. Additionally, if a split is achieved, parent classifiers of all nodes along the way to the root node are modified.

The concept of feature sharing is also explored through shape-based hierarchical compositional models [8], [9], [10]. These models are used for object categorization [8], [10], but also for multi-view detection, [9]. Different object categories can share parts or appearance. Parts on lower hierarchy levels are combined into larger parts on higher levels. In general, parts on lower levels are shared amongst various object categories, while those in higher levels are more category specific. This is similar to approach employed in this paper, however, in this paper, the feature sharing is obtained through a single-level support vector sharing.

The approach presented in [1] describes a road sign detection and recognition system based on sharing features. The detection subsystem is a two stage process comprised of color-based segmentation and SVM shape detection. The recognition subsystem comprises GentleBoost algorithm and Rotation-Scale-Translation Invariant template matching.

## III.  PROBLEM DEFINITION

In this paper, we focus on detection and classification of ideogram-based road signs. The class of all road signs exhibits various foreground variations with respect to the sign shape but also on presence or absence of thick red rim and ideogram type. We aim to jointly train detectors to discriminate road signs from backgrounds as well as to produce foreground variation estimates, i.e. subclass labels. We will use terms "foreground variation" and "subclass" interchangeably in the rest of the paper denoting the same concept. Fig. 1 depicts within-class road sign subclasses which we aim to estimate.

The class of all road signs is comprised out of five variations denoted with label $v$ from $S = \{1, 2, 3, 4, 5\}$. Subclass $v=1$ includes triangular warning signs with thick red. A small subset of the subclass members is shown in Fig. 1, i.e. the subclass contains nine different road signs. These signs belong to the category A according to the Vienna Convention [11]. Subclass $v=2$ contains circular "End of no overtaking zone" sign which belongs to the category C of informative signs. On the other hand, members of subclass $v=3$ "Priority road" and "End of Priority Road" are rhomb-shaped. Subclass $v=4$ includes square-shaped signs which are a subset of informative category C road signs. Finally, the subclass $v=5$ conatins circular "Speed Limit" signs characterized with the thick red rim which belongs to the category B of prohibitory signs. We discuss the described road-sign variations in our dataset and the motivation for our approach as follows.

First, we discuss motivation for partitioning road signs into subclasses according to the distribution shown in Fig. 1. The dataset is extracted from the video recorded with camera mounted on top of a moving vehicle. Video sequences are recorded at daytime, at different weather conditions [2]. Further, as Fig. 1 shows, the distribution of signs in the dataset is unbalanced, i.e. certain variations like triangular signs are characterized with large number of instances, while some others have a small number of occurrences. In particular, the "End of Priority Road" sign shown as a part of the subclass $v=3$ in Fig. 1 has only nine instances in training dataset. In the approach where we build a single detector for a particular subclass, it is clear that detector trained with only nine samples would have very poor detection rate. However, the "End of Priority Road" sign shares the same shape as the "Priority Road" sign. If we were to group them into a single subclass, we could exploit foreground-variation feature sharing. Other heterogeneous subclasses are designed with the same motivation. Note that the subclass $v=5$ is also a heterogeneous subclass, i.e. it contains various speed limit signs which share the thick red rim and the zero digit, since the speed limits are usually multiples of ten.

Second, according to the subclasses defined in Fig. 1, observe that signs belonging to different subclasses also share similarities. For example, the "End of no overtaking zone" (subclass $v=2$) sign and "End of Priority Road" (subclass $v=3$) both share the same distinctive crossover mark. This similarity could improve discrimination capability of both signs with respect to the background class. This suggests that if we solve detection and classification problem for all subclasses together, we could benefit from within-class feature sharing.

Therefore, due to the described characteristics of the dataset distribution, as well as the nature of sign similarities, we decided to employ a method presented in [3] where a classification function is learned jointly for all within-class variations. The aim of this approach is to form the classification function which could exploit the fact that different variations share features against backgrounds, but at the same time provide within-class discrimination.

## IV. DETECTION AND CLASSIFICATION APPROACH

The overall road sign detection and classification process is shown in Fig. 2. The detailed description is as follows.

For a given feature vector x ∈ $\mathbb{R}^n$ computed for an image patch, the goal is to decide whether it represents an instance of a road sign class and, if so, to produce the corresponding subclass estimate $v$ from $S = \{1, 2, 3, 4, 5\}$. Let $x_i$ denote feature vector of the $i$-th road sign training sample belonging to the subclass with label $v_i$. The feature vectors are given as HOG features [15]. According to the above defined parameters, the classification function $C(x,i)$ is defined as follows:

$$C(x,i) \begin{cases} > 0, \ x \text{ is foreground from the same subclass as } x_i \\ \leq 0, \text{ otherwise} \end{cases} \quad (1)$$

This corresponds to the non-parametric approach presented in [3]. The parametric approach [3] is simpler, however, it requires each subclass $v$ to be described with a specific parameter. The role of parameter is to describe members of the subclass in a unique way. In multiview domain, the parameter typically corresponds to the view angle or object pose. However, in the road sign domain the subclasses are heterogeneous and cannot be described with a single parameter. For this reason, the classification function employs the foreground sample feature vector $x_i$ in order to describe a specific subclass. Since the road sign subclasses are designed with the goal that signs within the subclass are similar, it is to be expected that they will also exhibit similar feature vector values $x_i$.

In order to satisfy requirements of within-class feature sharing against the background class, as well as within-class discrimination, the classification function $C(x,i)$ is represented as a product of two kernels:

$$C(x,i) = \sum_{s \in SV} \alpha_s \cdot k_\theta(x_s, x_i) \cdot k_x(x_s, x) \quad (2)$$

Parameter $\alpha_s$ denotes the Lagrange multiplier of the $s$-th support vector [12], $x_s$ the particular support vector, $x_i$ the $i$-th foreground training sample, while $k_\theta(x_s, x_i)$ denotes the within-class kernel and $k_x(x_s, x_i)$ the between-class kernel. In addition, the product of first two terms in equation (2) can be summarized into a single term $\alpha_s'(i)$, which denotes the weight of the $s$-th support vector $x_s$ for foreground sample $x_i$:

$$\alpha_s'(i) = \alpha_s \cdot k_\theta(x_s, x_i) \quad (3)$$

As a result, the support vectors for which the within-class kernel $k_\theta$ yields higher values will have a larger influence on the classification function. In this way, we achieve the within-class feature sharing as well as the within-class discrimination.

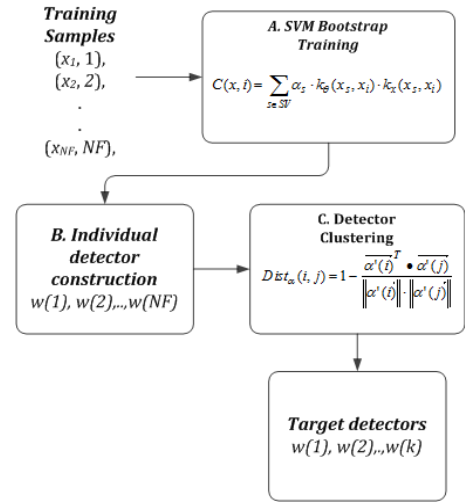

Fig. 2. Training and detector construction outline

### A. Classification Function Training

The classification function (2) training is achieved using SVM. The training samples take form of tuples $(x, i)$. Each foreground training sample $x$ is assigned its corresponding sample index $i$. Background training samples $x$ are obtained from image patches without road signs. Each background training sample $x$ can be associated with any index of a foreground training sample in order to form a valid tuple. More specifically, background samples $x$ is a negative with respect to all foreground samples. The number of such combinations is huge and corresponds to

$$\#(NB) \cdot \#(NF) \quad (4)$$

The parameter $\#(NB)$ corresponds to the total number of backgrounds and $\#(NF)$ to the total number of foreground training samples. Due to combinatorial complexity, including all negative samples in SVM training would not be practical. Therefore, the bootstrap training is employed as a hard negative mining technique. In bootstrap training, only $\#(NB)$ negatives are initially included in training. These samples are assigned foreground sample indices in a random fashion. After each training round, all negative samples are evaluated by the classification function (2). False positives are added to the negative set and the SVM training is repeated. This is an iterative process which converges when there are no more false positives to add.

### B. Individual Detector Construction

$C(x,i)$ is learned as a function of a foreground variation parameter $i$, rather than learning separate detectors for each $i$. Individual detectors $w(x,i)$ are obtained from the classification function by plugging in specific foreground sample values $x_i$ into (2) and (3)

$$w(x,i) = \sum_{s \in SV} \alpha_s'(i) \cdot k_x(x_s, x) \quad (5)$$

Note that with fixed foreground variation $i$ and known set of support vectors $x_s$, we can precompute the within-class kernel values $k_\theta(x_s, x_i)$ and consequently support vector weights $\alpha_s'(i)$. Therefore, the within-class kernel $k_\theta$ is not evaluated at detection time. Rather, at detection time, we only evaluate the between-class kernel $k_x$. This fact affects our choice for within-class and between-class kernels.

First, we discuss within-class kernel $k_\theta$. Road sign subclasses are difficult to separate and is therefore important for $k_\theta$ to be able to separate nonlinear problems. Therefore, Gaussian RBF kernel was chosen for that purpose

$$k_\theta(x, x_i) = \exp(-\eta \cdot D(x, x_i)) \quad (6)$$

where $D(x,x_i)$ denotes Euclidian distance and $\eta$ corresponding parameter. Due to the fact that RBF is evaluated only during training and detector construction, this choice doesn't impose performance penalty during detection.

Secondly, since the between-class kernel $k_x$ is evaluated during detection, it is important for $k_x$ to be fast. Therefore, we chose linear kernel for that purpose. By substituting the linear kernel formulation $k_x(x,x_i) = x_i^T x$ into (5) we obtain the final form of our detectors:

$$w(x,i) = \sum_{s \in SV} \alpha_s'(\theta) \cdot (x_s^T \cdot x) = w(i) \cdot x \quad (7)$$

In this way, the detection is achieved by applying a simple dot product between the image patch and the detector weights denoted as $w(i)$. Note that all detectors share the same set of support vectors. In this way, feature sharing among various detectors is achieved.

*C. Detection Approach*

In the detection process, we employ the well known sliding window technique. Each window is evaluated by a family of linear detectors $w(x,i)$ constructed according to (7). From all detector responses, the one with maximum value is chosen as a result. If this value is positive, the window is classified as a road sign belonging to the subclass of $x_i$. Otherwise, the window is discarded as a background. Note that this is similar to the $k$-nearest neighbors ($k$-NN) method, with parameter $k=1$, i.e. the object is simply assigned to the class of the nearest neighbor selected among all detectors [13].

However, in the *1*-NN approach, the number of evaluated detectors is significant, i.e. corresponds to the number of foreground samples #(*NF*). Evaluating all #(*NF*) detectors at the detection stage would make the detection extremely slow. In addition, since foreground samples belonging to the same subclass are similar, there may be redundancy among detectors. In order to identify a representative set from a family of total #(*NF*) detectors, we use the $k$-medoids clustering technique. The $k$-medoids technique is chosen due to its simplicity and also because it is less prone to outlier influence than, for instance, $k$-means method. Clustering yields a set of $k < $ #(*NF*) medoids which are then used in the detection phase. In clustering, each detector $w(x, i)$ is represented with a vector of its support vector weights:

$$\overrightarrow{\alpha'(i)} = \alpha_1'(i), \alpha_2'(i), \ldots, \alpha_{SV}'(i) \quad (8)$$

where $\alpha_s'(i)$, $s \in \{1,\ldots, SV\}$ denotes the particular support weight defined with (3), while $SV$ denotes total number of support vectors. As a distance measure, we used $Dist_\alpha(i, j)$ defined as follows:

$$Dist_\alpha(i, j) = 1 - \frac{\overrightarrow{\alpha'(i)}^T \bullet \overrightarrow{\alpha'(j)}}{\left\|\overrightarrow{\alpha'(i)}\right\| \cdot \left\|\overrightarrow{\alpha'(j)}\right\|} \quad (9)$$

The appropriate number of medoids $k$ is chosen in an iterative process, where we decrease number of medoids gradually and measure the clustering quality. Initially, the target number of medoids, i.e. cluster centers $k$ is set to 50% of the initial number of detectors, i.e. foreground samples. This number was chosen by a rule of thumb. Then, we apply clustering according to the chosen number of medoids $k$. In order to measure clustering quality, we compute corresponding silhouette value [14]. The silhouette value provides an estimate of how the well obtained medoids represent the data in their corresponding clusters. In each iteration, the target number of clusters is decreased by a certain factor and the above process is repeated. Final clustering outline is chosen as the one which yields the best silhouette value.

V. EXPERIMENTAL RESULTS

In this section, we describe the evaluation of the above described method according to the foreground variation distribution presented in Section III. In our experiments, we used three disjoint datasets, described below [2].

As we already stressed, the distribution shown in Fig. 1 is unbalanced, i.e. subclass $v=1$ contains at least three times more samples that other classes. In such unbalanced datasets, there is a possibility for specific road sign variation with smaller number of samples to be treated as a noise, e.g. subclass $v=3$. In order to test this hypothesis, we experimented with the number of samples per class. Namely, we observed detection performance for $N_{POS}$=200, 300 and 400 samples per subclass. For the subclasses with the number of samples lower than $N_{POS}$, we simply use all the samples for the subclass, e.g. subclasses $v=2$, $v=3$ and $N_{POS}$=400. The training samples are extracted randomly from training dataset pool comprising 2153 road signs. Negative dataset contains 4000 image patches extracted randomly from images of background outdoor scenes. In order to monitor clustering performance, we use the test dataset comprised out of 3000 cropped images. Details of the test dataset are given in Section V-B.

*A. Training Approach*

The training of detectors is achieved as described in Section IV. As features, we used HOG vectors [15] computed from training images cropped and scaled to 24 x 24 pixels. Cell size is set to 4 pixels, where each cell is normalized within a block of four cells. In order to increase performance, we use block overlapping with block stride set to size of single cell. The training is achieved using SVMlight [16] with multiplicative kernel. In contrast to [3], where parameter $\eta$ of the RBF kernel

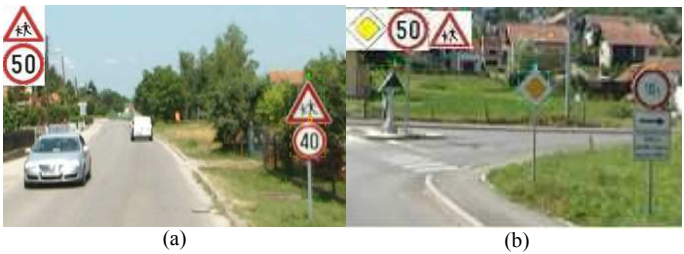

Fig. 3. Examples of a detection and classification: (a) correct classification, (b) correct classification and false positive.

(6) is set to a fixed value, we perform cross validation on training dataset in each bootstrapping round in order to obtain the best value. We compared both training approaches and the one with optimized η value yields a lower number of support vectors. This suggests a better mapping in the transformed feature space. More specifically, with training set comprised out of 1325 positive samples and 4000 negative samples, the training with the fixed η value yields a set of 857 support vectors. On the other hand, the training with the optimized η decreases the number of support vectors for 10%.

The training yields a family of #(NF) detectors. The #(NF) corresponds to the number of foreground training samples which depends on the number of training samples per subclass $N_{POS}$, Table I. Due to performance reasons, we use $k$-medoids clustering in order to select a representative set of detectors from total #(NF) detectors. The clustering is implemented in Matlab according to the Partitioning Around Medoids method [17]. In each clustering iteration, we monitor the silhouette value and the validation results on the test dataset comprised out of cropped images. Note that this dataset is disjoint from test dataset used for detection. Interestingly, better silhouette values correspond to a smaller number false negatives obtained from validation on test data. The results of clustering are summarized in Table I. depending on the $N_{POS}$ and #(NF). Resulting number of detectors $k$ corresponds approximately to 30% of total training samples #(NF). Lower $k$ values lead to poor validation results and small silhouette values.

*B. Detection and Classification Results*

The test dataset for detection evaluation contains 1038 images in 720x576 resolution. From the total 1038 images, we selected 200 images and used them for performance evaluation. In these images, there were 214 physical road signs.

Table II. shows the detection and foreground estimation results for the three case studies. We report the detection rate D, the classification rate C, the false positive rate FP and the false positive rate per image FP/I. D is defined as a number of detected signs with respect to the total number of signs, while C and FP correspond to the number of correct classifications and false detections with respect to the total number of signs, respectively. FP/I corresponds to the number of false detections with respect to the number of images. Columns denoted with Δ sign show differences in above metrics with respect to configuration denoted by $N_{POS}$=200.

We started the experiment with $N_{POS}$=200 samples per subclass. This configuration achieves overall detection rate of 90% with false positive rate of 43%. Next, $N_{POS}$=300 achieves 4% rise in detection rate giving total 94%, and 3% rise in classification rate, i.e. 93%. However, it is characterized with a large false positive rate of 55%. In sliding window detection, some computer vision libraries like OpenCV [18] employ false positive detection policy where an object must exhibit at least $n$ detections in order to be accounted as a result. This is understandable, since sliding window technique exhibits multiple responses around single object. In this work, we didn't experiment with this property, however we believe that it would decrease false positive rate. Finally, the configuration $N_{POS}$=400 exhibits worse results with respect to $N_{POS}$=300. This supports our hypothesis that the unbalanced dataset $N_{POS}$=400 treats subclasses with a lower number of samples as a noise giving the lower overall detection rate.

Table III. depicts D and C rates, as well as the FP rate per specific subclass $v$ for configurations $N_{POS}$=300 and $N_{POS}$=200, respectively. The subclass $v$=1 achieves better results when a larger number of samples is used. This is understandable, since this subclass comprises a large number of different road signs. Interestingly, the subclass $v$=3 which has only 98 samples (8% of the total samples for subclass $v$=1) achieves detection and classification rate of 100% in all case studies. Subclasses $v$=2 and $v$=4 achieve lower detection and classification rates with respect to other subclasses. Subclass $v$=2 is circle-shaped but lacks red rim in order to share features with subclass $v$=5. On the other hand, subclass $v$=4 is rectangle-shaped and gains less benefit from feature sharing with other subclasses. Subclass $v$=5 obtains similar results for $N_{POS}$=300 and $N_{POS}$=200. FP distribution per subclass for $N_{POS}$=200 shows that subclass $v$=4 exhibits the largest number of false positives, i.e. 58%. Examples of false positives classified as members of subclass $v$=4 include building windows. On the other hand, $N_{POS}$=300 yields a rather balanced FP distribution, where subclasses $v$=1, 4 and 5 obtain FP rate of approximately 30%. Examples of detection and classification are given in Fig. 3a and Fig. 3b. Fig. 3a illustrates an example of a correct classification, where "Speed Limit" sign is classified as a member of subclass $v$=5 (orange dotted line), while the "Children" sign is classified as a member of subclass $v$=1 (green dotted line). Fig. 3.b illustrates correct classification, as well as two false positives. The "Priority Road" sign is correctly classified as a subclass $v$=3 (cyan dotted line). The "Weight Limit" sign was not present in training data, however, due to similarity with "Speed Limit" sign, it is classified as a member of subclass $v$=5 (orange dotted line). The latter one indicates within-class feature sharing. The triangle-like object is incorrectly classified as a member of subclass $v$=1 (green dotted line).

VI. CONCLUSION

In this paper, we considered a road-sign detection technique based on a multiplicative kernel. One of the major challenges was a poorly balanced dataset, where triangular warning signs have at least three times more instances than other subclasses. Our approach is based on a premise that different sign subclasses share features which discriminate them from backgrounds. Therefore, instead of learning a dedicated detector for each subclass, we trained single classification function for all subclasses using SVM [3]. Individual detectors are afterwards constructed from a shared set of support vectors. Major benefit of such approach with respect to separately trained detectors lies in feature sharing which enhances the detection rate for subclasses with lower number of samples.

TABLE I.    CLUSTERING RESULTS

| N_POS | #(NF) | k |
|---|---|---|
| 200 | 833 | 288 |
| 300 | 1132 | 332 |
| 400 | 1325 | 388 |

TABLE II.    DETECTION AND CLASSIFICATION

| $N_{POS}$ | D | ΔD | C | ΔC | FP | ΔFP | FP/I | ΔFP/I |
|---|---|---|---|---|---|---|---|---|
| 200 | 91% | - | 90% | - | 45% | - | 47% | - |
| 300 | 94% | +3% | 93% | +3% | 55% | +10% | 58% | +11% |
| 400 | 90% | -1% | 90% | 0% | 43% | -2% | 45% | -3% |

TABLE III.    RESULTS PER SUBCLASS

| | $N_{POS}$ = 300 | | | $N_{POS}$ = 200 | | |
|---|---|---|---|---|---|---|
| v | D | C | FP | D | C | FP |
| 1 | 100% | 100% | 34% | 91% | 91% | 15% |
| 2 | 82% | 77% | 0% | 82% | 82% | 0% |
| 3 | 100% | 100% | 4% | 100% | 100% | 5% |
| 4 | 88% | 82% | 31% | 76% | 71% | 58% |
| 5 | 92% | 92% | 31% | 93% | 93% | 22% |

In comparison to partition based approaches, this approach does not require the training samples to be labeled with subclass parameters in order to learn classification function. This fact has proven to be useful for our domain, since the road sign subclasses are heterogeneous and it is hard to describe a subclass with a single parameter. Instead, each training sample is labeled with its corresponding HOG feature vector. In this way we obtain #(NF) subclass parameters and consequently #(NF) detectors from the classification function, where #(NF) corresponds to the number of foreground samples. However, due to performance issues, we perform clustering in order to obtain a representative set of detectors from a set of total #(NF) detectors. The reduced set of detectors is used in detection.

Using the described method, we achieved the best detection rate of 94% at a relatively high false positive rate of 55%. We experimented with different numbers of samples per subclass in order to observe the effect on detection rate. The obtained results showed that detectors trained on a limited number of samples, i.e. 300 samples per class obtain better detection results with respect to the larger number of samples. Due to the fact that this method has shown promising results in road sign domain, in future work we plan to explore its applicability in the domain of multiview vehicle detection.

ACKNOWLEDGMENT

The authors wish to thank Josip Krapac for useful suggestions on early versions of this paper.